\documentclass[11pt]{article}

\usepackage[preprint]{acl}

\usepackage{times}
\usepackage{latexsym}

\usepackage[T1]{fontenc}
\usepackage{multirow}

\usepackage[utf8]{inputenc}
\usepackage{enumitem}   
\usepackage{amssymb}

\usepackage{microtype}
\usepackage{makecell}
\usepackage{subcaption}

\usepackage{inconsolata}
\usepackage{booktabs}

\usepackage{graphicx}
\usepackage{amsmath}
\usepackage{booktabs}
\usepackage{multirow}
\usepackage{array}
\usepackage{xcolor}
\usepackage{colortbl}
\usepackage{arydshln}    
\usepackage{ragged2e}

\usepackage{cleveref}

\usepackage[normalem]{ulem}

\definecolor{myblue}{rgb}{0.933, 0.933, 0.996}
\definecolor{myred}{rgb}{0.996, 0.933, 0.933}

\definecolor{ChargeColor}{HTML}{E8F0FE}    
\definecolor{ImpColor}{HTML}{FFF4E5}       
\definecolor{FineColor}{HTML}{E8F5E9}      

\definecolor{FactColor}{RGB}{113, 109, 145}
\definecolor{InferColor}{RGB}{187, 121, 147}
\definecolor{ResultColor}{RGB}{141, 170, 196}
\definecolor{OtherColor}{RGB}{248, 246, 235}

\definecolor{TagColor}{RGB}{30, 80, 160}
\definecolor{FactColor}{RGB}{225, 235, 250}
\definecolor{InferColor}{RGB}{230, 245, 230}
\definecolor{ResultColor}{RGB}{252, 235, 220}
\definecolor{OtherColor}{RGB}{240, 240, 240}

\usepackage{tcolorbox}
\usepackage{xcolor}

\title{HKJudge: A Legal Discourse-Annotated Corpus for Interpreting \\What Courts Find, How They Reason, and What They Rule}

\author{
  Xi Xuan$^{1}$, \;
  Wenxin Zhang$^{2}$, \;
  Yufei Zhou$^{1}$, \;
  King-kui Sin$^{1}$, \;
  Chunyu Kit$^{1}$ \\
  $^{1}$City University of Hong Kong \quad
  $^{2}$University of Chinese Academy of Sciences \\
  \texttt{\{xixuan3, ctckit\}@cityu.edu.hk}
}

\begin{document}
\maketitle
\begin{abstract}

Court judgments are central to legal practice and jurisprudence, yet discourse analysis of Hong Kong judgments has received limited attention, owing largely to the absence of expert-annotated corpora. We introduce the Hong Kong Judgment Discourse Dataset (HKJudge), the first sentence-level expert-annotated legal discourse corpus. HKJudge includes criminal judgments across all five levels of HK's court hierarchy, comprising $\sim$290k sentences and $\sim$6.5 million tokens, fully annotated by legal linguistics experts. We design a two-tier discourse schema that captures what facts a court finds, how it reasons, and what it rules. At the sentence level, each sentence is assigned one of 26 rhetorical roles. At the span level, sentences are further annotated with three sentencing elements (charge, imprisonment term, fine). Ten legal linguistics annotators produced the annotations with an inter-annotator agreement of $\kappa = 0.8$. We formulate two tasks on HKJudge, termed rhetorical role classification and legal element extraction, and provide the first benchmark evaluation of four BERT-based models, two open-source LLMs under zero-shot and fine-tuning settings, and four commercial LLMs on both tasks. Our work demonstrates the value of sentence-level discourse annotation for modeling the structure of HK judgments and provides a rich data foundation for future work on legal judgment prediction. The HKJudge dataset and code are available at\footnote{https://github.com/xuanxixi/HKJudge}.

\end{abstract}
\vspace{-0.3cm}
\section{Introduction}
Court judgments are among the most important legal genres for the legal 
profession, in both legal practice and jurisprudence~\citep{cheng2008discursive}. 
They are performative speech acts whose fundamental function is to 
adjudicate, and they simultaneously serve declaratory, justificatory, and 
legitimating purposes within a single document~\citep{ 
maley2014language}. Making judgments tractable for downstream NLP tasks, including legal 
search~\citep{werner1981corporation, mo2025survey}, case analysis~\citep{li2025legalagentbench}, and legal judgment prediction 
(LJP)~\citep{gillman2001s,he2024agentscourt,dancy2026ai}, requires modeling knowledge of the generic 
structures of legal documents. Such structural modeling reduces search space, facilitates the identification of rhetorical segments, thereby facilitating the working efficiency of court judgments ~\citep{saravanan2010identification,han2018structural,kalamkar2022corpus}.

While substantial progress has been made in modeling court judgment structure for Indian \citep{ghosh2019identification, kalamkar2022corpus, nigam2025legalseg}, European \citep{rosas2007european,held2026lacour}, United States \citep{robinson2013structure, williams2022jurisdiction, shu2024lawllm} and Chinese mainland \citep{xiao2018cail2018,liebman2020mass,fei2025internlm} jurisdictions, comparable resources for Hong Kong (HK) case law remain scarce. HK court judgments, produced within a bilingual common-law jurisdiction with its own appellate hierarchy \citep{cheng2008discursive,yu2023negotiation,xuan2024efficient}, follow drafting conventions and discourse structures that differ from those of the corpora discussed above, particularly in citation practice \citep{cheng2015moral}, sentencing discourse \citep{yu2025linguistic,xuan2026disentangling}, and bilingual reasoning \citep{cheng2016revisiting,xuan25_spsc}. Direct transfer of models trained on other jurisdictions is therefore unreliable.

\begin{figure*}[t]
  \centering
  \includegraphics[width=0.96\textwidth]{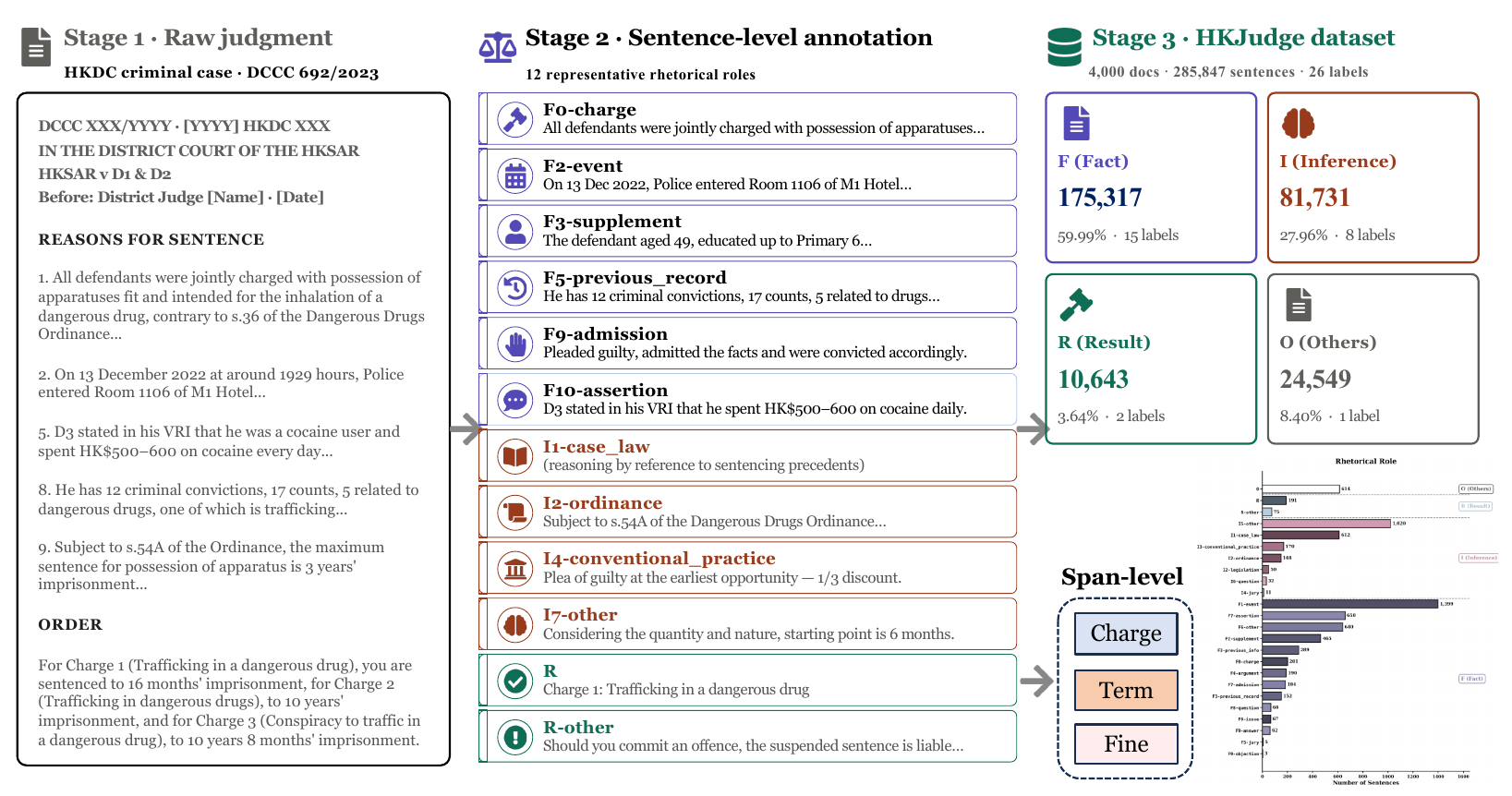}
\caption{Overview of the HKJudge annotation process.
Stage~1 (left): an anonymized Hong Kong District Court criminal judgment.
Stage~2 (center): each sentence is labeled with one of 26 rhetorical roles (see Appendix~\ref{sec:schema} for full definitions), grouped into four categories: Fact (F), Inference (I), Result (R), and Others (O); twelve representative labels are shown.
Stage~3 (right): summary of the HKJudge dataset (4{,}000 documents, $\sim$ 290k sentences, 26 labels) and three span-level element types: Charge, Term, and Fine, extracted from R-tagged sentences.}
  \label{fig:hkjudge_pipeline}
  \vspace{-0.4cm}
\end{figure*}

Previous research in this domain has highlighted the importance of 
annotated datasets for training effective models. However, currently 
there is no publicly available dataset for the HK JLP task that is 
\textit{fully annotated by legal linguistics experts}. Many existing studies have relied on relatively small annotated datasets with only coarse-grained, three-level labels of \textit{facts}, \textit{reasoning}, and \textit{ruling}, so that charges and prior records share the same label, and case-law reasoning is not distinguished from that citing an Ordinance, limiting their effectiveness for LJP systems in real-world scenarios. The few existing HK legal dataset resources each have important limitations. For instance, the Legal-NLP Dataset of \citep{sen2023analyzing} was constructed from HKLII judgments using regular expressions and semantic parsing, without expert annotation of rhetorical structure. The HKCFA Judgment 97-22 dataset \citep{xuan2026trans} targets legal translation rather than discourse analysis, and covers only Court of Final Appeal judgments. The LegalHK dataset \citep{shi2025legalreasoner}, from the LegalReasoner framework, relies on GPT-4 to extract structured information from judicial documents with \textit{only partial manual review by judicial experts}, excludes appellate cases from the court of appeal and the court of final appeal, and captures only three coarse functional blocks without sentence-level rhetorical roles.

We see the need to introduce a unified mode of study
that can quickly incorporate new areas and applications
of law.  In this work, we \textit{develop a uniform discourse
schema for characterising a HK court judgment. Discourse analysis}, the study of how texts are organized into functional segments above the sentence level ~\citep{gill2000discourse, joty2019discourse, gee2025introduction}, has been successfully applied to areas like news events~\citep{nakshatri2025talking}, dialogue understanding~\citep{ko2023discourse,xuan2026wstx},
web documents~\citep{liu2023webdp}, legal documents~\citep{sovrano2025discolqa}, and synthetic-content characterization~\citep{xuan2024conformer, 11434679, 10890034, 11462804, 11461768, ZHANG2026132741}. In legal domain, ~\citep{sovrano2025discolqa} effectively use discourse analysis for legal question answering,
improving state-of-the-art without fine-tuning or re-training the language models
on the regulations at hand.

In this work, we develop a legal discourse schema to address this
need. At its core, our schema seeks to answer three questions
about each judgment: (1) \textit{what} facts the court finds,
(2) \textit{how} it reasons, and (3) \textit{what} it rules.
We show that both pretrained encoders and LLMs struggle to model
this schema, whereas legal linguistics experts label it with high
inter-annotator agreement. In sum, this paper makes three key contributions:
\vspace{-0.1cm}
\begin{enumerate}[leftmargin=*,itemsep=2pt,topsep=2pt]
    \item \textbf{Introducing, Annotating and Modeling a Legal Discourse Schema:}
    We develop a HK legal discourse schema, consisting of 3 span-level and 26 sentence-level rhetorical role classes, some of which are shown in Figure~\ref{fig:hkjudge_pipeline}. We construct court judgments dataset annotated by legal linguistics experts, with 148,600 spans and 292,240 rhetorical role annotations. We show that our schema can be labeled with high inter-annotator agreement. Additionally, we show LLM models (zero-shot and fine-tuned) outperform BERT-based models.

    \item \textbf{Web Scraping Public HK Case Law:}
We build web scrapers and collect over 4{,}000 judgments spanning 1968--2024 from five Hong Kong courts, namely the Court of Final Appeal, the Court of Appeal, the Court of First Instance, the District Court, and the Magistrates' Courts. Hong Kong judgments are subject to HKSAR Government copyright but publicly available for private and academic use.\footnote{\url{https://www.judiciary.hk/en/other_information/disclaimer.html}.} Our scrapers comply with the access policies of the Judiciary's Legal Reference System\footnote{\url{https://legalref.judiciary.hk/}} and use rate-limited requests.

    \item \textbf{Benchmarking BERT-based and LLM-based Methods on Court Judgments Annotation:}
We evaluate four BERT-based methods, open-source LLMs, and commercial LLMs (including GPT-4, Claude, and Gemini) under zero-shot and fine-tuned settings. Although fine-tuning yields substantial gains, all LLMs still fall noticeably short of human expert annotators and commercial LLMs, highlighting the value of expert annotation and pointing to open challenges in legal LLM reasoning.
\end{enumerate}

\section{A Legal Discourse Schema}
\label{sec:schema1}

Court judgments serve multiple functions, including
adjudication, declaration, justification, and
legitimation~\citep{maley2014language}. Modeling their structure
at the discourse level therefore provides an effective entry
point into legal reasoning~\citep{carlsonbuilding, prasad2017penn},
and supports downstream tasks including legal
search~\citep{mo2025survey}, case
summarization~\citep{li2025legalagentbench}, and legal judgment
prediction~\citep{aletras2016predicting, malik2021ildc,
shi2025legalreasoner, dancy2026ai}. Hong Kong court judgments in
particular can be segmented along the rhetorical roles of
heading, introduction, facts, analysis, and
conclusion~\citep{cheng2008contrastive}.

As shown in Figure~\ref{fig:hkjudge_pipeline}, modeling the 
different rhetorical roles of a legal doctrine as 
\textit{discourse units} and how they interact can be an 
effective way of discerning 
meaning~\citep{carlsonbuilding,prasad2017penn}. Identifying these parts poses a basic test of a model's legal
reasoning and unlocks practical applications, as
\citet{hendrycks1cuad} demonstrated in contract review. We
accordingly introduce a schema that captures this distinction
at two levels, starting with sentence-level annotations and
extending to span-level extraction of three result elements,
termed \textit{charge}, the offence of conviction;
\textit{imprisonment term}, the length of custodial sentence;
and \textit{fine}, the monetary penalty.

\subsection{Discourse-level Schema}
\label{sec:sentence-schema}

The 4 discourse functions we identify in our legal discourse  schema are
\colorbox{FactColor}{\textcolor{black}{\textbf{Fact}}},
\colorbox{InferColor}{\textcolor{black}{\textbf{Inference}}},
\colorbox{ResultColor}{\textcolor{black}{\textbf{Result}}},
and \colorbox{OtherColor}{\textcolor{black}{\textbf{Other}}}. The first three functions correspond to
the three layers of information carried by every judgment,
whereas Other is a residual class covering procedural or
formulaic sentences that fall outside the preceding three. The
first three functions, Fact, Inference, and Result, capture how
a judgment proceeds from established facts, through the court's
reasoning, to the ruling, inspired by the contrastive analysis of
HK court judgment structure done
by~\citet{cheng2008contrastive}.
We describe each category in turn.

\begin{itemize}
  \item A \textbf{\colorbox{FactColor}{\textcolor{black}{Fact (F)}}}
        sentence typically reports information presented to
        the court without expressing the court's own
        evaluation. We distinguish 15 sub-tags by procedural
        origin and evidentiary status: F0-charge, F1-issue,
        F2-event, F3-supplement, F4-previous\_info,
        F5-previous\_record, F6-argument, F7-jury, F8-other,
        F9-admission, F10-assertion, F11-question, F12-answer,
        F13-objection, and F14-instructions2jury. F11--F14
        generally originate inside the courtroom, while
        F2--F5 generally originate outside it. (e.g.\
        \emph{``The defendant is 24 and has 2 conviction
        records, which include 2 `Theft' offences, 3 `Robbery'
        offences and 1 `Attempted Robbery' offence.''} is
        tagged F3-supplement.)

  \item An \textbf{\colorbox{InferColor}{\textcolor{black}{Inference (I)}}}
        sentence is one in which the court itself reasons
        toward its decision. We distinguish 8 sub-tags by the
        authority appealed to: I1-case\_law, I2-ordinance,
        I3-legislation, I4-conventional\_practice, I5-jury,
        I6-assertion, I7-other, and I8-question. The boundary
        with Fact tends to rest on whose voice is speaking,
        since a party's argument and a judge's reasoning can
        be lexically similar. (e.g.\ \emph{``It does not
        identify a purpose which it thinks would be beneficial
        and then construe the statute to fit it.''} is tagged
        I1-case\_law.)

  \item A \textbf{\colorbox{ResultColor}{\textcolor{black}{Result (R)}}}
        sentence states the disposition of the case, under
        two sub-tags: R, the final determination, and R-other,
        supplementary content attached to it (clarifications,
        calculations, statements of consequence). The boundary
        can be subtle, since explanatory material may
        intervene between successive operative rulings within
        a single paragraph. (e.g.\ \emph{``For that reason,
        and in light of the concession made by the appellant
        in relation to the third respondent, the appeal must
        be dismissed against all of the respondents.''} is
        tagged R.)
\end{itemize}

\noindent We give full definitions of the rhetorical role sub-tags in
Appendix~\ref{sec:schema}. Judgments rendered by the court of
appeal and the court of final appeal embed the lower court's
Facts, Inferences, and Results into their own text, tagged
F4-previous\_info. Some of these sentences retain a secondary
discourse function such as I8-question, and we allow both tags
in such cases.

\subsection{Span-Level Schema}
\label{sec:span-schema}

We define 3 span-level element types during our annotation
process, applied to sentences tagged Result. These cover what the court decides, termed \colorbox{ChargeColor}{\uline{\textbf{\textit{charge}}}} denoting the offence of conviction; \colorbox{ImpColor}{\uline{\textbf{\textit{imprisonment term}}}} denoting the length of custodial sentence; and \colorbox{FineColor}{\uline{\textbf{\textit{fine}}}} denoting the monetary penalty.
The three types are mutually exclusive at the span level, with
each span identifying one sentencing decision. Charges and their penalties often appear in the same R sentence, producing spans of more than one type. We do not annotate spans for other
sentencing options such as suspended sentences, community
service orders, or disqualification orders, since these surface
infrequently in our data; the containing sentence is retained
under R at the sentence level.

\section{Dataset Construction}

In this section, we describe how we operationalized the schema
discussed in Section~\ref{sec:schema1}. We scrape a dataset of
Hong Kong criminal case laws from 1968 to 2024 across five
court levels, which we discuss in Section~\ref{sec:source}. We build an
annotation framework, described in
Section~\ref{sec:annotation}, and enlist ten annotators, who
collectively annotate $\sim$290k law sentences.

\subsection{Dataset Source and Web Scraping}
\label{sec:source}

The corpus contains over 4,000 Hong Kong criminal judgments from 1968 to 2024, spanning all five court levels with criminal jurisdiction. We scrape these judgments from the Hong Kong
Legal Information Institute
(HKLII),\footnote{\url{https://www.hklii.hk/databases}} a
public-access platform that aggregates court judgments released
by the Hong Kong Judiciary. Raw judgments data has a mix of court of final appeal judgments
(20\%), court of appeal judgments (20\%), court of first instance
judgments (20\%), district court judgments (20\%) and
magistrates' courts judgments (20\%). We audit the HKLII output against
the Judiciary's Legal Reference System
(LRS),\footnote{\url{https://legalref.judiciary.hk/}} and find
judgments that HKLII does not cover, has not updated, or
renders as image-based PDFs with high OCR error
rates;\footnote{For example, older judgments from
\textsc{HKMagC} on HKLII are stored as scanned PDFs; we extract
text directly from the LRS PDF instead.} for these we fall
back to LRS PDFs read through
\texttt{pdfplumber}.\footnote{\url{https://github.com/jsvine/pdfplumber}}

Hong Kong judgments are subject to HKSAR Government copyright
but are publicly available for private and academic
use.\footnote{\url{https://www.judiciary.hk/en/other_information/disclaimer.html}}
Although HKLII and LRS websites are publicly accessible, they
employ a range of mechanisms (e.g.\ timeouts, dynamically
generated URLs, cookie-based access) that make them difficult
to scrape. To circumvent
these, our scrapers are robust and mimic human web-browsing
behavior. We develop a generalized scraper for Hong Kong court
judgment public-access websites using
scrapy\footnote{\url{https://scrapy.org/}} and
selenium-webdriver.\footnote{\url{https://www.selenium.dev/}}
In order to scrape HKLII, we launch three Google Compute Engine
(GCE) instances for a total of 40 compute
hours.\footnote{We will release our code for scraping with
Docker images created to perform these scrapes. Given the
difficulty in creating this dataset, we believe these routines
constitute a considerable resource for academic inquiries into
Hong Kong case law.}

\subsection{Annotation Process}
\label{sec:annotation}
We recruited 10 annotators from a pool of 30 annotator
candidates, all graduate students majoring in legal
linguistics, selected on the basis of their strong academic
backgrounds and familiarity with legal processes. We then
trained all of the annotators for multiple rounds, until they
were achieving above an 80\% accuracy in both discourse and
span identification tasks, based on a gold-label set that we
constructed. After reaching this agreement level, we began
accepting completed tasks from annotators. We had multiple
rounds of conferencing throughout the period of annotation
where we discussed edge-cases, and maintained a WeChat channel
throughout the annotation process that was continually
monitored. The annotation process spanned from October 2025 to
May 2026. Together, the annotators annotated $\sim$290k sentences,
with a 10\% overlap, from which we calculated a $\kappa = .8$.

We found that our annotators could learn to identify different discourse and span levels in most contexts quite easily. Appendix~\ref{sec:schema_exp} presents an example of legal linguistics expert annotation on a Court of Final Appeal judgment (FACC 22/2018).
However, most of the error and ambiguity of the annotation
process derived from when to distinguish F9-admission from
F10-assertion (e.g.\ ``\textit{Mr.\ WU accepted that the
defendant had caused PW1 to lose his properties, but the
defendant did not retain any}'' can be read either as a
counsel's neutral assertion or as an admission on behalf of
the defendant). The decision usually depends on many factors,
e.g.\ whether the speaker has authority to bind the defendant.
Despite many rounds of training, annotators still sometimes struggled with borderline cases; in such circumstances, they consulted senior legal linguistics experts for adjudication.

\begin{figure}[t]
    \centering
    \includegraphics[width=0.95\columnwidth]{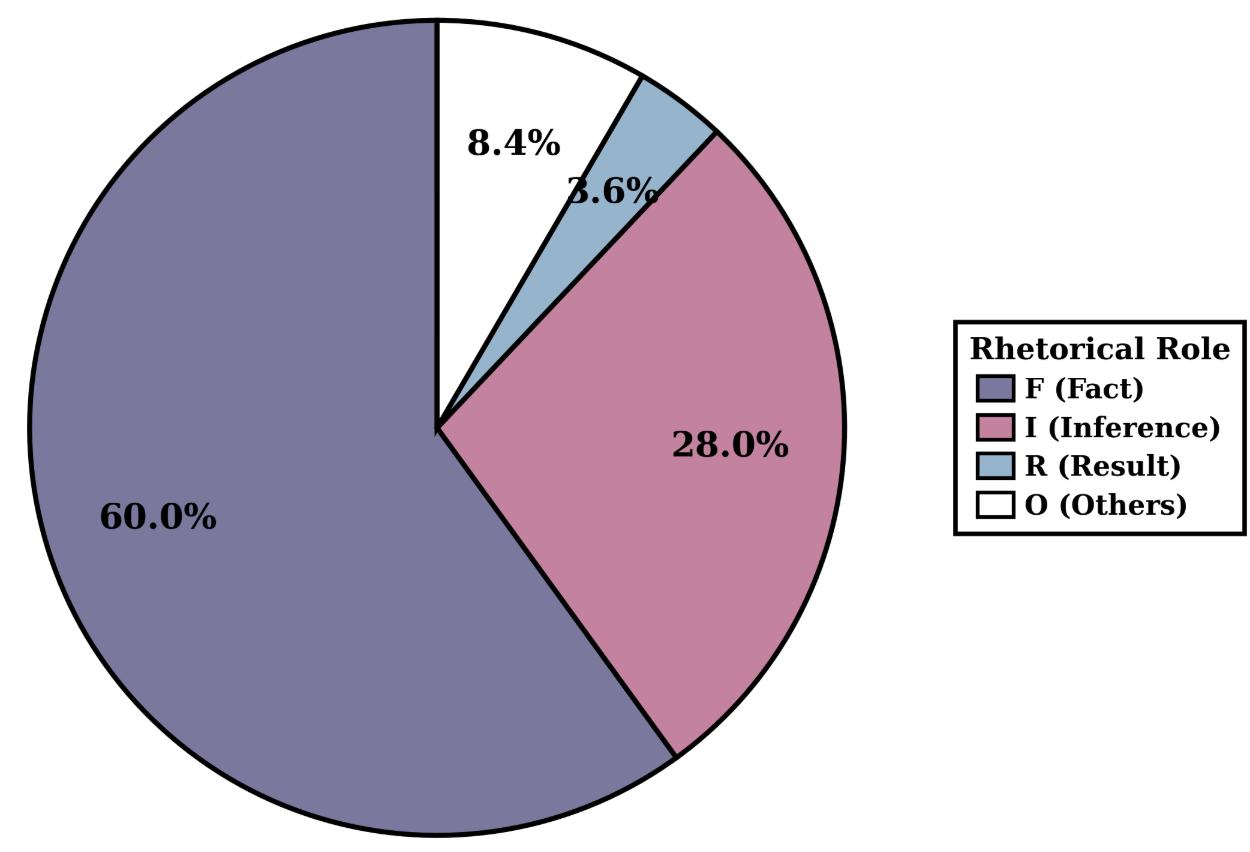}
    \caption{\small Distribution of Rhetorical Roles within the HKJudge Dataset.}
    \label{fig:category_overview}
    \vspace{-0.7cm}
\end{figure}

\subsection{Dataset Description and Statistics}
\begin{figure}[t]
    \centering
    \includegraphics[width=\columnwidth]{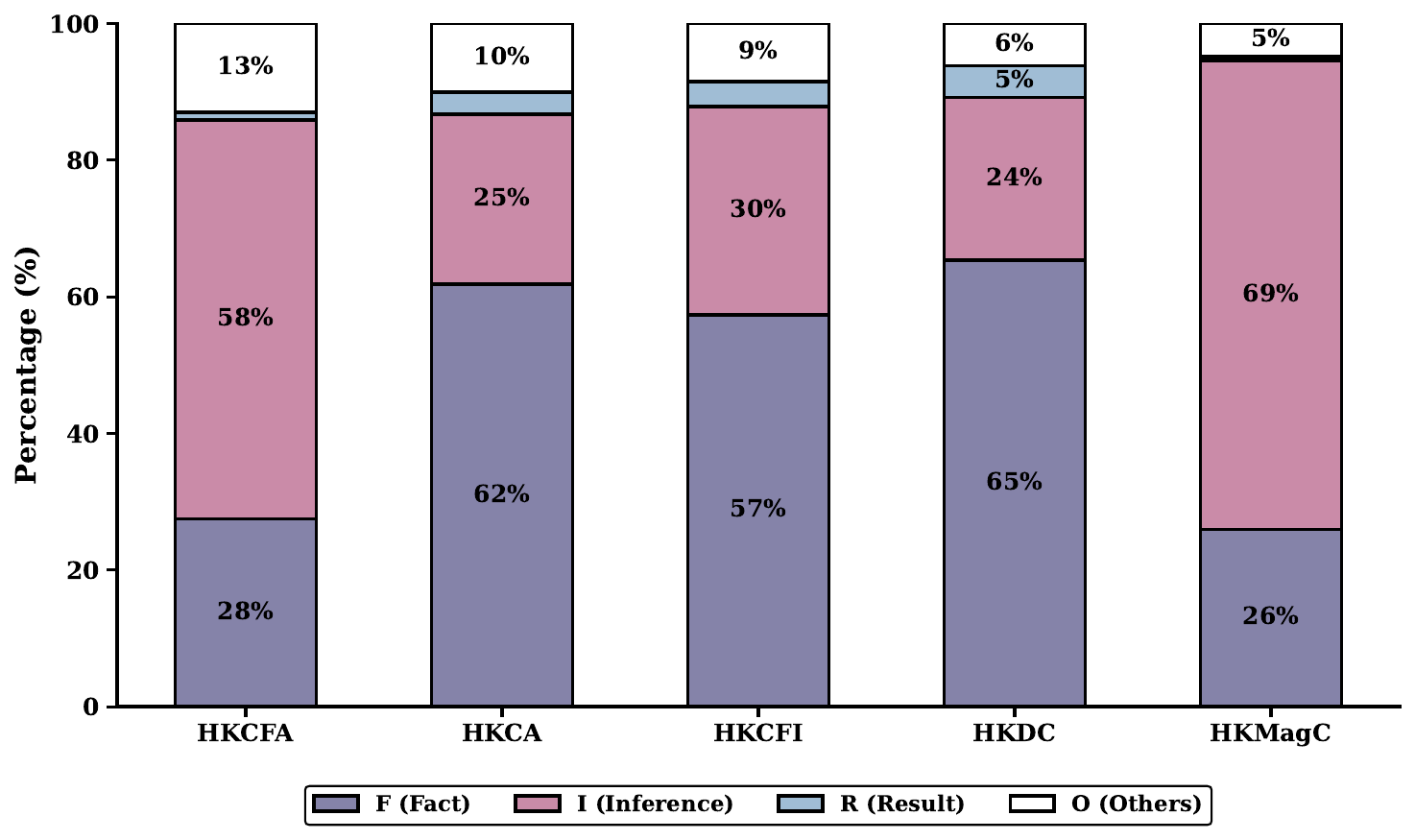}
    \caption{Discourse function distribution across five court levels. Higher courts (HKCFA, HKCA) allocate a larger share to \textit{Inference}, matching their emphasis on legal reasoning and precedent.}
    \label{fig:court_x_category}
\end{figure}

\begin{figure}[t]
    \centering
    \includegraphics[width=\columnwidth]{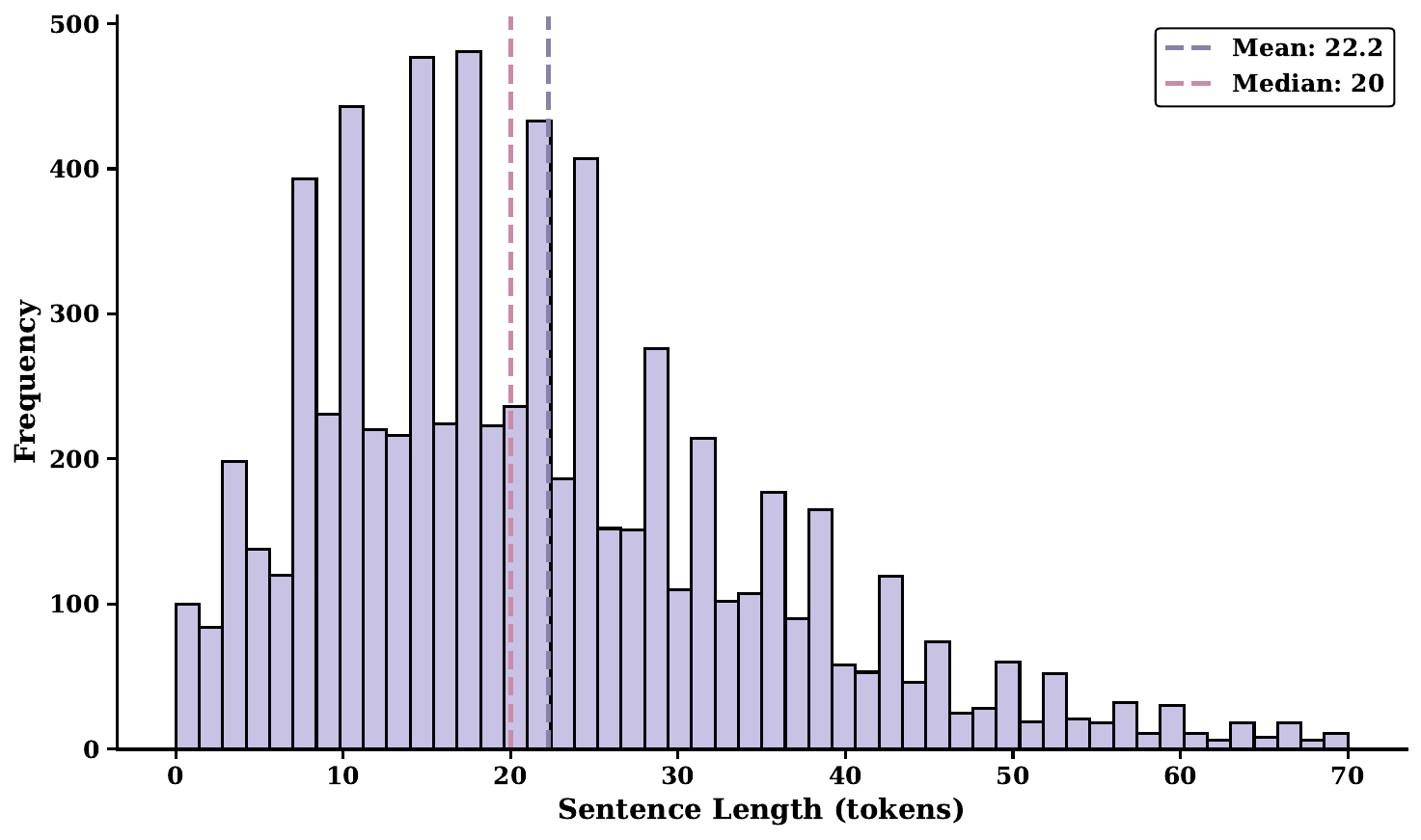}
    \caption{Distribution of annotated sentence lengths in HKJudge Dataset. Mean (purple dashed) and median (pink dashed) are indicated.}
    \label{fig:sentence_length}
\end{figure}

As shown in Table~\ref{tab:dataset_stats}, the court judgments
we annotate average $1{,}631.9$ tokens and $73.1$ sentences per
document. The judgments we focused on are criminal cases; see Appendix~\ref{app:judgment_sample} for an HKCFI Judgment Example (Case No. HCCC 12/2021). The HKJudge corpus contains $4{,}000$ documents, $292{,}240$ annotated
sentences (of which $285{,}847$ are unique), and $6{,}527{,}600$ tokens.
Multi-tagged sentences account for 1.97\% of the corpus. As shown in Figures~\ref{fig:category_overview}, \ref{fig:court_x_category}, and \ref{fig:sentence_length}, sentence-level
annotations are distributed across different rhetorical roles,
termed \textit{fact} accounting for 59.99\% of all
annotations, \textit{inference} for 27.96\%, \textit{others}
for 8.40\%, and \textit{result} for 3.64\%. The HKJudge dataset is available at\footnote{https://github.com/xuanxixi/HKJudge}.

\begin{table}[t]
  \centering
  \small
  \setlength{\tabcolsep}{3pt}
  
  \begin{tabular}{p{4.0cm}r@{\hspace{1.0cm}}r@{\hspace{1.0cm}}r}
    \Xhline{1.4pt}
    \multicolumn{4}{c}{\textbf{HKJudge Dataset Overall Statistics}} \\
    \hline
    Documents                       & & \multicolumn{2}{r}{4{,}000} \\
    Unique sentences                & & \multicolumn{2}{r}{285{,}847} \\
    Sentence--tag pairs             & & \multicolumn{2}{r}{292{,}240} \\
    Total tokens                    & & \multicolumn{2}{r}{6{,}527{,}600} \\
    Avg.\ sentences per document    & & \multicolumn{2}{r}{73.1} \\
    Avg.\ tokens per document       & & \multicolumn{2}{r}{1{,}631.9} \\
    Avg.\ tokens per sentence       & & \multicolumn{2}{r}{22.3} \\
    Multi-labeled sentences         & & \multicolumn{2}{r}{5{,}757 (1.97\%)} \\
    \Xhline{1.4pt}
  \end{tabular}
  
  \vspace{0.3em}

  \begin{subtable}{\columnwidth}
    \centering
    \begin{tabular}{lrrr}
      \Xhline{1.4pt}
      \multicolumn{4}{c}{\textbf{Distribution Across Hong Kong Courts}} \\
      \hline
      \textbf{Court} & \textbf{\# Docs} & \textbf{\# Sents} & \textbf{\# Tokens} \\
      \hline
      Court of Final Appeal     & 800 & 56{,}243 & 1{,}256{,}187 \\
      Court of Appeal           & 800 & 59{,}418 & 1{,}325{,}962 \\
      Court of First Instance   & 800 & 58{,}791 & 1{,}312{,}854 \\
      District Court            & 800 & 57{,}873 & 1{,}293{,}716 \\
      Magistrates' Courts       & 800 & 59{,}915 & 1{,}338{,}881 \\
      \Xhline{1.0pt}
      \multicolumn{4}{c}{\textbf{Discourse-level Distribution}} \\
      \hline
      \textbf{Category} & \textbf{\# Sents} & \textbf{Pct. (\%)} & \textbf{\# Tokens} \\
      \hline
      F (Fact)        & 175{,}317 & 59.99 & 3{,}848{,}629 \\
      I (Inference)   &  81{,}731 & 27.96 & 2{,}168{,}542 \\
      R (Result)      &  10{,}643 &  3.64 &   212{,}758 \\
      O (Others)      &  24{,}549 &  8.40 &   297{,}671 \\
      \Xhline{1.4pt}
    \end{tabular}
    
  \end{subtable}
  
  \caption{Dataset statistics for the HKJudge dataset.}
  \label{tab:dataset_stats}
\end{table}

\section{Legal Discourse and Entity Modeling}
\label{sec:tasks}

We frame two tasks using the data we 
collect: \textit{Rhetorical Role Classification} and \textit{Element 
Extraction}. Each sentence in a judgment document is 
labeled with one of four top-level categories: 
Fact (F), Inference (I), Result (R), or Other (O). 
T1 is a sentence classification task that assigns 
each F, I, or O sentence to one or more 
sub-categories from our annotation 
scheme~(\S\ref{sec:schema}). T2 is a generative 
extraction task that prompts large language 
models to identify charges, imprisonment terms, 
and fines from R sentences. We will first describe these 
tasks, then discuss methods, with a particular 
focus on how we use this setup to interrogate the 
reasoning capabilities of large language models.

\subsection{Task Formulation}
\label{subsec:formulation}

F and I sentences describe \textit{what} happened 
and \textit{how} the court reasoned. R sentences 
state \textit{what} the court ruled, including 
charges, imprisonment terms, and fines. O 
sentences are residual. We 
model F, I, and O as classification, and R as element extraction.

\paragraph{Task 1: Rhetorical Role Classification.}
The goal of this task is to develop models capable of performing semantic segmentation on court judgments by identifying and classifying rhetorical roles (RR). Let $C = \{c_1, c_2, \ldots, c_n\}$ represent a collection of court judgments, where $c_i \in C$ consists of a sequence of sentences $S_i = \{s_{i1}, s_{i2}, \ldots, s_{im}\}$, with $m$ representing the number of sentences in court judgment $c_i$. The task is to assign a rhetorical role label $y_{ij} \in Y$
to each sentence $s_{ij}$, where $Y$ is the predefined set of
26 rhetorical role labels defined in
Appendix~\ref{sec:schema}, organized into four top-level
categories: Fact (F), Inference (I), Result (R), and Other
(O).
Formally, the task can be described as:
\begin{equation}
  f: S_i \rightarrow Y
\end{equation}
\noindent where $Y$ is defined as:
\begin{equation}
  Y = Y_F \cup Y_I \cup Y_R \cup Y_O
\end{equation}
\noindent where $f$ is a function that maps each sentence $s_{ij}$ in a judgment $c_i$ to its corresponding rhetorical role label $y_{ij}$. Thus, the goal is to find:
\begin{equation}
  f(s_{ij}) = y_{ij}, \quad \forall s_{ij} \in S_i, \quad y_{ij} \in Y
\end{equation}
The input to the system is a court judgment $c_i$, and the output is rhetorical role labels corresponding to each sentence in the court judgment:
\begin{equation}
  f(S_i) = \{y_{i1}, y_{i2}, \ldots, y_{im}\}, \quad y_{ij} \in Y
\end{equation}
We benchmark various large language models using accuracy and macro-F1 scores.
\paragraph{Task 2: Legal Element Extraction.}
Given a judgment sentence $s_i$ labeled as R, we extract three element types defined by the Hong Kong sentencing framework~\citep{young2017sentencing,xue2024leec}: charge ($\mathsf{Charge}$), imprisonment term ($\mathsf{Term}$), and fine ($\mathsf{Fine}$). The task output is formalized as:
\begin{equation}
    E_i = f(s_i) = \{(t_j, v_j)\}_{j=1}^{k_i},
\end{equation}
where $f(\cdot)$ denotes the extraction function implemented by a large language model, $t_j \in \mathcal{T} = \{\mathsf{Charge}, \mathsf{Term}, \mathsf{Fine}\}$ indicates the legal element type, and $v_j$ is the textual span extracted from $s_i$. The cardinality $k_i$ varies across sentences, as a single sentence may convey multiple charges or penalties, or contain no extractable element ($E_i = \emptyset$). 
We also benchmark large language models on element extraction using precision and macro-F1 scores. We count a prediction $(t_j, v_j)$ as correct if its type matches the gold type and its span shares at least 80\% of tokens with the gold span (after removing stop words and punctuation) with length no more than twice the gold span.

\begin{table*}[t]
  \centering

  \small
  \setlength{\tabcolsep}{3pt}
  \resizebox{\textwidth}{!}{%
  \begin{tabular}{l cccc cccc}
    \Xhline{1.4pt}
    \multirow{2}{*}{\textbf{Model}} & \multicolumn{4}{c}{\textbf{Rhetorical Role Classification}} & \multicolumn{4}{c}{\textbf{Legal Element Extraction}} \\
    \cmidrule(lr){2-5} \cmidrule(lr){6-9}
    & \textbf{Accuracy} & \textbf{AUC} & \textbf{Precision} & \textbf{Macro-F1} & \textbf{Accuracy} & \textbf{AUC} & \textbf{Precision} & \textbf{Macro-F1} \\
    \Xhline{1.0pt}
    \multicolumn{9}{c}{\textit{BERT-based methods}} \\
    \Xhline{0.6pt}
    \rowcolor{myblue} LegalBERT          & \underline{64.235}$_{\pm1.793}$ & \underline{71.926}$_{\pm1.581}$ & \underline{61.547}$_{\pm1.864}$ & \underline{61.924}$_{\pm1.832}$ & \underline{58.314}$_{\pm1.987}$ & \underline{66.082}$_{\pm1.843}$ & \underline{55.471}$_{\pm2.052}$ & \underline{56.243}$_{\pm2.011}$ \\
    NeuralJudge        & 63.521$_{\pm1.842}$ & 71.043$_{\pm1.625}$ & 60.874$_{\pm1.913}$ & 61.273$_{\pm1.876}$ & 57.482$_{\pm2.041}$ & 65.317$_{\pm1.892}$ & 54.628$_{\pm2.103}$ & 55.391$_{\pm2.054}$ \\
    \rowcolor{myred}  ML-LJP             & \textbf{64.923}$_{\pm1.762}$ & \textbf{72.583}$_{\pm1.548}$ & \textbf{61.832}$_{\pm1.821}$ & \textbf{62.213}$_{\pm1.789}$ & \textbf{58.832}$_{\pm1.954}$ & \textbf{66.728}$_{\pm1.812}$ & \textbf{55.741}$_{\pm2.018}$ & \textbf{56.521}$_{\pm1.976}$ \\
    JurBERT            & 63.913$_{\pm1.815}$ & 71.624$_{\pm1.602}$ & 61.218$_{\pm1.887}$ & 61.634$_{\pm1.854}$ & 57.962$_{\pm2.014}$ & 65.731$_{\pm1.871}$ & 55.142$_{\pm2.078}$ & 55.924$_{\pm2.035}$ \\
    \Xhline{0.6pt}
    \multicolumn{9}{c}{\textit{Commercial LLMs}} \\
    \Xhline{0.6pt}
    GPT-4              & 73.532$_{\pm1.487}$ & 78.214$_{\pm1.348}$ & 70.583$_{\pm1.524}$ & 70.921$_{\pm1.498}$ & 68.421$_{\pm1.687}$ & 73.582$_{\pm1.612}$ & 66.518$_{\pm1.712}$ & 66.832$_{\pm1.684}$ \\
    Claude-3.5-Sonnet  & 73.804$_{\pm1.475}$ & 78.421$_{\pm1.341}$ & 70.612$_{\pm1.518}$ & 70.931$_{\pm1.493}$ & 68.742$_{\pm1.672}$ & 73.821$_{\pm1.598}$ & 67.521$_{\pm1.702}$ & 67.831$_{\pm1.674}$ \\
    \rowcolor{myred}  Claude-Opus-4      & \textbf{77.152}$_{\pm1.328}$ & \textbf{81.532}$_{\pm1.214}$ & \underline{71.832}$_{\pm1.412}$ & \underline{72.134}$_{\pm1.385}$ & \textbf{72.031}$_{\pm1.524}$ & \textbf{76.842}$_{\pm1.452}$ & \underline{68.072}$_{\pm1.564}$ & \underline{68.354}$_{\pm1.538}$ \\
    \rowcolor{myblue} Gemini-2.5-Pro     & \underline{76.842}$_{\pm1.342}$ & \underline{81.218}$_{\pm1.228}$ & \textbf{72.143}$_{\pm1.398}$ & \textbf{72.421}$_{\pm1.372}$ & \underline{71.823}$_{\pm1.538}$ & \underline{76.524}$_{\pm1.468}$ & \textbf{68.342}$_{\pm1.552}$ & \textbf{68.621}$_{\pm1.524}$ \\
    \Xhline{1.4pt}
  \end{tabular}%
  }
  \caption{Performance of BERT-based methods and commercial LLMs on rhetorical role classification and legal element extraction. \textbf{Bold} numbers indicate the best score and \underline{underlined} numbers represent the second best within each category. Red and blue rows highlight the best and second-best models in each group, respectively.}
    \label{tab:results1}
  \vspace{-0.3cm}
\end{table*}

\subsection{Baselines}

We conduct experiments on both BERT-based and
LLM-based methods. For BERT-based methods:

\begin{itemize}
    \item \textbf{LegalBERT}~\citep{chalkidis2020legal} pre-trains BERT on legal documents from scratch.
    \item \textbf{NeuralJudge}~\citep{yue2021neurjudge} enhances pre-trained BERT with LJP-specific fine-tuning.

    \item \textbf{ML-LJP}~\citep{liu2023ml} integrates contrastive learning and Graph Attention Networks to model law article interactions.

    \item \textbf{JurBERT}~\citep{masala2024improving} extends LegalBERT with a Sliding Encoder for improved long-context understanding.

\end{itemize}

For LLM-based methods, we compare both open-source and commercial LLMs:

\begin{itemize}
    \item \textbf{LLaMA 3.1}~\citep{grattafiori2024llama} and \textbf{Qwen 2.5}~\citep{hui2024qwen2} represent state-of-the-art open-source language models.
    
    \item \textbf{GPT 4}~\citep{sanderson2023gpt}, \textbf{Claude 3.5 Sonnet}~\citep{benzon2025miriam}, \textbf{Claude Opus 4}~\citep{joshi2026architectural}, and \textbf{Gemini 2.5 Pro}~\citep{comanici2025gemini} demonstrate strong performance as proprietary models.
\end{itemize}

\begin{table*}[t]
  \centering
  
  \small
  \setlength{\tabcolsep}{3pt}
  \resizebox{\textwidth}{!}{%
  \begin{tabular}{l cccc cccc}
    \Xhline{1.4pt}
    \multirow{2}{*}{\textbf{Model}} & \multicolumn{4}{c}{\textbf{Rhetorical Role Classification}} & \multicolumn{4}{c}{\textbf{Legal Element Extraction}} \\
    \cmidrule(lr){2-5} \cmidrule(lr){6-9}
    & \textbf{Accuracy} & \textbf{AUC} & \textbf{Precision} & \textbf{Macro-F1} & \textbf{Accuracy} & \textbf{AUC} & \textbf{Precision} & \textbf{Macro-F1} \\
    \Xhline{1.0pt}
    LLaMA-3.1-8B               & 61.594$_{\pm2.137}$ & 67.281$_{\pm1.942}$ & 58.328$_{\pm2.112}$ & 58.621$_{\pm2.084}$ & 56.218$_{\pm2.283}$ & 62.142$_{\pm2.118}$ & 53.421$_{\pm2.293}$ & 53.742$_{\pm2.261}$ \\
    \quad + Fine-tuning         & 64.473$_{\pm1.876}$ & 71.421$_{\pm1.728}$ & 61.231$_{\pm1.884}$ & 61.521$_{\pm1.857}$ & 60.274$_{\pm2.065}$ & 67.418$_{\pm1.887}$ & 57.312$_{\pm2.048}$ & 57.642$_{\pm2.018}$ \\
    \rowcolor{myblue} LLaMA-3.1-70B              & 70.231$_{\pm1.624}$ & 74.582$_{\pm1.483}$ & 67.428$_{\pm1.687}$ & 67.752$_{\pm1.658}$ & 65.962$_{\pm1.812}$ & 70.231$_{\pm1.738}$ & 63.142$_{\pm1.852}$ & 63.421$_{\pm1.825}$ \\
    \rowcolor{myred} \quad + Fine-tuning         & 72.371$_{\pm1.547}$ & 76.318$_{\pm1.412}$ & 68.423$_{\pm1.612}$ & 68.741$_{\pm1.583}$ & 68.213$_{\pm1.728}$ & 72.487$_{\pm1.658}$ & 64.218$_{\pm1.781}$ & 64.542$_{\pm1.753}$ \\
    \hdashline
    Qwen-2.5-7B                & 66.014$_{\pm1.823}$ & 70.341$_{\pm1.714}$ & 62.918$_{\pm1.887}$ & 63.291$_{\pm1.854}$ & 61.812$_{\pm1.948}$ & 66.218$_{\pm1.842}$ & 58.871$_{\pm2.014}$ & 59.218$_{\pm1.983}$ \\
    \quad + Fine-tuning         & 71.423$_{\pm1.612}$ & 73.821$_{\pm1.527}$ & 64.412$_{\pm1.658}$ & 64.702$_{\pm1.628}$ & 67.321$_{\pm1.752}$ & 70.142$_{\pm1.687}$ & 60.218$_{\pm1.814}$ & 60.541$_{\pm1.785}$ \\
    Qwen-2.5-14B               & 69.842$_{\pm1.687}$ & 73.421$_{\pm1.563}$ & 66.987$_{\pm1.724}$ & 67.302$_{\pm1.695}$ & 65.421$_{\pm1.842}$ & 69.318$_{\pm1.768}$ & 62.918$_{\pm1.876}$ & 63.241$_{\pm1.848}$ \\
    \quad + Fine-tuning         & 71.742$_{\pm1.605}$ & 75.143$_{\pm1.487}$ & 66.873$_{\pm1.642}$ & 67.213$_{\pm1.614}$ & 68.582$_{\pm1.718}$ & 71.842$_{\pm1.652}$ & 62.821$_{\pm1.781}$ & 63.142$_{\pm1.753}$ \\
    \rowcolor{myblue} Qwen-2.5-72B               & \underline{70.752}$_{\pm1.612}$ & \underline{75.218}$_{\pm1.478}$ & \underline{69.218}$_{\pm1.628}$ & \underline{69.547}$_{\pm1.598}$ & \underline{66.521}$_{\pm1.798}$ & \underline{70.842}$_{\pm1.724}$ & \underline{65.083}$_{\pm1.812}$ & \underline{65.421}$_{\pm1.785}$ \\
    \rowcolor{myred}  \quad + Fine-tuning         & \textbf{72.987}$_{\pm1.524}$ & \textbf{76.831}$_{\pm1.402}$ & \textbf{69.348}$_{\pm1.587}$ & \textbf{69.682}$_{\pm1.562}$ & \textbf{68.823}$_{\pm1.687}$ & \textbf{72.918}$_{\pm1.614}$ & \textbf{65.214}$_{\pm1.752}$ & \textbf{65.531}$_{\pm1.728}$ \\
    \Xhline{1.4pt}
  \end{tabular}%
  }
  \caption{Performance of open-source LLMs (LLaMA-3.1 and Qwen-2.5 series) under both zero-shot and fine-tuning training. \textbf{Bold} numbers indicate the best score and \underline{underlined} numbers represent the second best across all models. Red and blue rows highlight the best and second-best results, respectively.}
  \label{tab:results2}
\end{table*}

\subsection{Evaluation Metrics}
To evaluate the performance of models, we adopt a set of standard metrics including Accuracy, AUC, Precision, and Macro-F1, which are commonly used in classification and element extraction tasks. For each sentence in the dataset, the predicted label (26 rhetorical roles for classification and 3 span-level element types for extraction) is considered correct if it matches the label assigned by the human expert annotator. To ensure statistical reliability, we report two times the standard deviation for all metrics using 1,000 bootstrap runs \cite{efron1986bootstrap} on the test dataset.

\subsection{Implementation Details}

For all BERT-based baselines and open-source LLMs, we fine-tune them on the HKJudge dataset, using the pre-segmented sentences of original court judgments as input and their rhetorical roles as labels. Open-source and commercial LLMs are additionally evaluated in the zero-shot setting without task-specific training. We fine-tune LLaMA-3.1 and Qwen-2.5 using the LLaMA-Factory framework~\citep{zheng2024llamafactory}, with the AdamW optimizer, learning rate $1 \times 10^{-5}$, weight decay 0.01, and cosine learning rate schedule. The number of training epochs is set to 3.

\section{Results and Analysis}
\vspace{-0.2cm}
In this section, we present the results of our experiments on rhetorical role classification and legal element extraction, and analyze the performance of the different models. Tables~\ref{tab:results1} and~\ref{tab:results2} summarize the evaluation metrics for BERT-based methods, commercial LLMs, and open-source LLMs under zero-shot and fine-tuning settings.
\vspace{-0.2cm}
\paragraph{Rhetorical Role Classification}
Among the evaluated BERT-based methods, ML-LJP attains the highest overall performance on rhetorical role classification, and the remaining encoder baselines, including LegalBERT, NeuralJudge, and JurBERT, follow closely with only marginal differences. The ability of ML-LJP to capture relationships between law articles through its multi-law-aware contrastive representation contributes to its performance, since rhetorical roles in legal documents are not assigned in isolation but follow conventional patterns of citation and reasoning that depend on the surrounding statutory context. 

In contrast, the open-source LLMs benefit substantially from scale. Fine-tuning LLMs further improves performance across all open-source variants, although the marginal gain from fine-tuning decreases as the base model grows larger. The Qwen-2.5-72B model with fine-tuning attains the strongest open-source result on rhetorical role classification, exceeding the ML-LJP by a clear margin, which highlights the advantage of large instruction-tuned decoders over encoder-only architectures for discourse-level legal tasks.

Among the commercial LLMs, Claude-Opus-4 leads on accuracy and AUC, while Gemini-2.5-Pro leads on precision and macro-F1. The gap between the strongest open-source model and the commercial systems is smaller than the gap between the BERT-family baselines and that strongest open-source model, which suggests that the principal bottleneck on rhetorical role classification is the reasoning supporting the tag decision rather than the decision itself, and that this reasoning capability scales with model capacity and instruction tuning.

\paragraph{Legal Element Extraction}

Performance decreases across most of the evaluated model families when moving from rhetorical role classification to legal element extraction, indicating that span-level extraction of charges, imprisonment terms, and fines is the more challenging task on HKJudge. The BERT-family ranking on extraction is largely preserved from the classification setting, with ML-LJP remaining the strongest in this group, although the absolute scores degrade more substantially than they do on classification. 

For the open-source LLMs, fine-tuning produces larger gains on extraction than on classification, with the improvements observed for LLaMA-3.1-8B and Qwen-2.5-72B among the largest single-task gains across our experiments. This is consistent with our hypothesis that extraction depends more heavily on task-specific supervision than classification does, since the surface conventions of HK sentencing spans, particularly the ordinance citation format and the phrasing of suspended and concurrent terms, are unlikely to be adequately represented in general instruction tuning. The commercial LLMs retain the lead on extraction, although their advantage over the strongest fine-tuned open-source model is reduced relative to the classification setting, which nonetheless demonstrates the potential of commercial LLMs for span-level legal reasoning tasks.

\section{Conclusion and Future Work}

\vspace{-0.2cm}
We presented \textbf{HKJudge}, the first sentence-level discourse corpus of Hong Kong court judgments fully annotated by legal linguistics experts. We benchmarked four BERT-based encoders, two open-source LLMs under zero-shot and fine-tuning settings, and four commercial LLMs on rhetorical role classification and legal element extraction. Across both tasks, performance increases monotonically from BERT-based encoders to fine-tuned open-source LLMs to commercial LLMs. ML-LJP achieves the highest scores among encoders, fine-tuned Qwen-2.5-72B leads the open-source LLMs, and Claude-Opus-4 and Gemini-2.5-Pro lead the commercial LLMs.

We highlight three findings from these results. First, the performance gap between the strongest open-source and commercial models is smaller than the gap between BERT-based baselines and the strongest open-source LLM, indicating that the principal bottleneck on rhetorical role classification is the legal reasoning that supports each tag assignment rather than the assignment itself, and that this reasoning capability scales with model size and instruction tuning. Second, fine-tuning yields larger gains on legal element extraction than on rhetorical role classification, consistent with the surface conventions of Hong Kong sentencing spans, including ordinance citation formats and the phrasing of suspended and concurrent terms, being underrepresented in general-purpose instruction tuning. Third, all evaluated models fall noticeably short of expert annotators, indicating open challenges for legal LLM reasoning over legal discourse.

For future work, we will use the HKJudge dataset proposed in this paper to explore and address legal judgment prediction in Hong Kong, a task that supports legal professionals (practitioners, law firms, judicial bodies, policymakers, and government departments), improving judicial efficiency and justice, and enabling citizens to anticipate case outcomes without costly legal consultation. Together with the dataset and benchmark released in this work, we hope HKJudge will serve as a step toward legal discourse modeling for LegalAI.
\section*{Limitations}
Our research focuses on Hong Kong criminal case law, which is governed by the common-law tradition and exhibits a bilingual drafting practice with highly standardized rhetorical conventions. Consequently, the discourse schema and trained models developed in this work may not be directly applicable to judgments from civil-law jurisdictions, monolingual common-law systems, or non-criminal legal areas such as civil and family proceedings. The results of our study, therefore, may not cover all countries or types of legal documents.

In addition, the boundary between certain Fact sub-categories (notably F9-admission vs.\ F10-assertion) remains subject to interpretive judgment by the annotator. Our span-level schema is also restricted to three sentencing elements (charge, imprisonment term, and fine), leaving alternative outcomes such as suspended sentences, community service orders, and disqualification orders for future extension.

\section*{Ethics Statement}

Our dataset and evaluation benchmark contain no personal, sensitive, or private information; they consist solely of publicly available data.

\section*{Acknowledgments}

The work described in this paper was fully supported by a grant from the Research Grants Council of HKSAR, China (Project No. CityU 11602524). We also thank the expert annotators in legal linguistics for their valuable contributions.


\vspace{-0.2cm}
\bibliography{custom}
\clearpage
\appendix
\twocolumn 
\onecolumn

\section{Use of AI Assistants}
We used Claude Opus 4.7 and Sonnet 4.6 for coding, shortening texts and editing LaTeX more efficiently.

\section{HKCFI Judgment Example (Case No. HCCC 12/2021)}
\label{app:judgment_sample}
\begin{figure}[!ht]
    \centering
    \setlength{\fboxrule}{1pt}
    \setlength{\fboxsep}{0pt}
    \fbox{\includegraphics[page=1, width=0.90\linewidth]{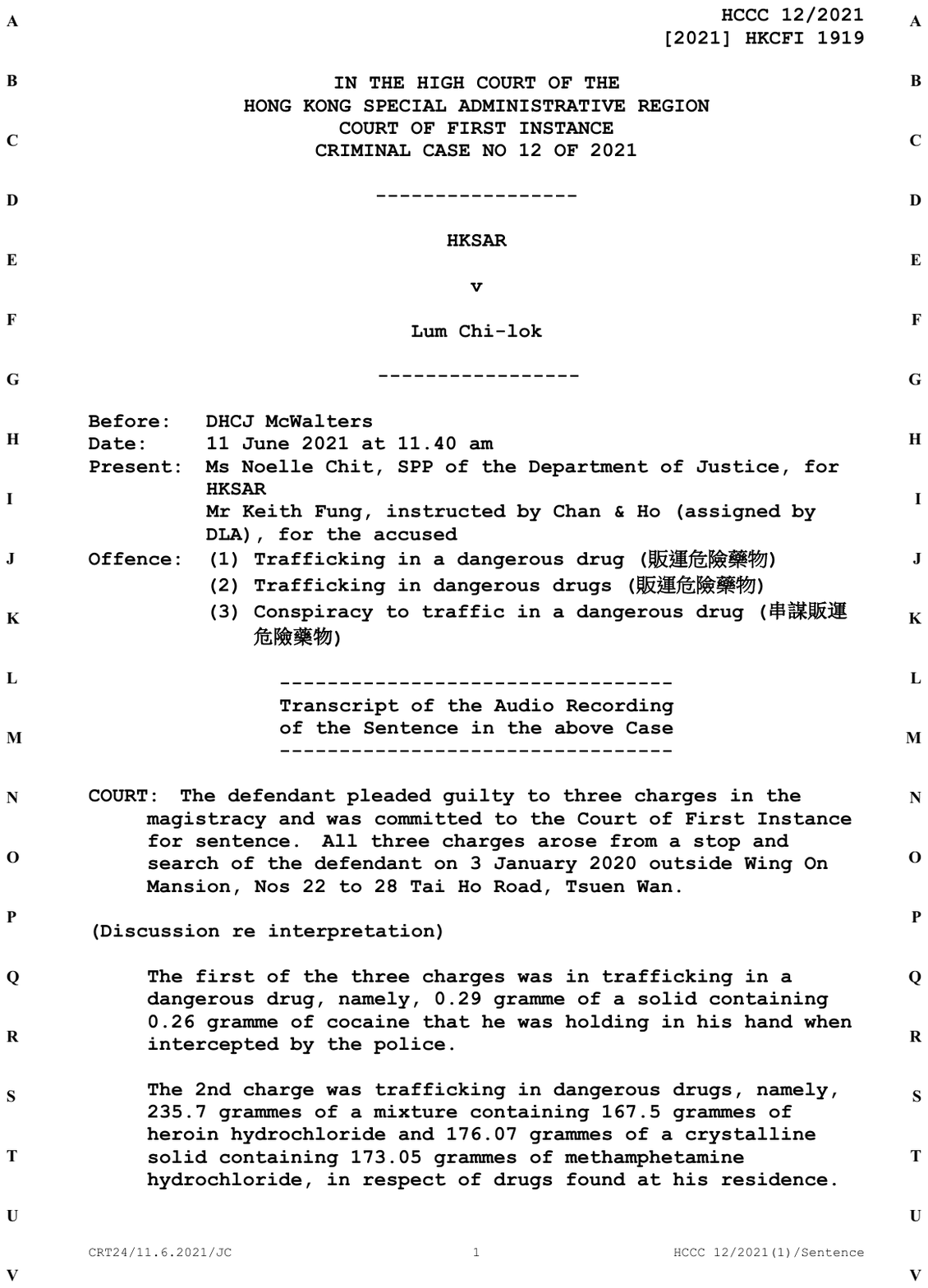}}
    \caption{Example of the first page of a Hong Kong Court of First Instance judgment (\textit{[2021] HKCFI 1919}, Case No.\ HCCC 12/2021).}
    \label{fig:hkcfi_judgment_sample}
\end{figure}

\section{The HKJudge Legal Discourse Annotation Scheme}
\label{sec:schema}

\begin{table*}[!ht]
\centering
\small
\renewcommand{\arraystretch}{1.25}
\setlength{\tabcolsep}{6pt}

\begin{tabular}{@{} >{\centering\arraybackslash}p{1.0cm}
                    >{\RaggedRight\arraybackslash}p{3.5cm}
                    >{\RaggedRight\arraybackslash}p{10.5cm} @{}}

\Xhline{1.4pt}
\textbf{Category} & \textbf{Rhetorical Role Tag} & \textbf{Description} \\
\Xhline{1.4pt}

\rowcolor{FactColor}
\multicolumn{3}{l}{\textbf{F (Fact)} \, --- \, \textit{What kinds of information are presented in a court hearing and/or recorded in a judgment?}} \\
\Xhline{1.4pt}

\textbf{F0} & \textbf{\textcolor{TagColor}{F0-charge}} & Charge(s) or offence(s) in a criminal case; a special case of \textbf{F1-issue}. \\
\textbf{F1} & \textbf{\textcolor{TagColor}{F1-issue}} & Marking the key issue(s) that a judgment is intended to judge. \\
\textbf{F2} & \textbf{\textcolor{TagColor}{F2-event}} & Descriptions of time, location, individuals involved, causes, processes, and outcomes of an event; constitutes the narrative of the core incident under consideration. \\
\textbf{F3} & \textbf{\textcolor{TagColor}{F3-supplement}} & Supplementary information regarding the core incident, such as background details of relevant individuals (typically in mitigation arguments) and events, attributes of objects, etc. \\
\textbf{F4} & \textbf{\textcolor{TagColor}{F4-previous\_info}} & Applicable \emph{exclusively to appeal cases}; citations or paraphrasing of prior court rulings or reasoning of the \emph{same case}. \\
\textbf{F5} & \textbf{\textcolor{TagColor}{F5-previous\_record}} & Historical records of verdict, convictions and/or sentence from previous \emph{unrelated} trial(s). \\
\textbf{F6} & \textbf{\textcolor{TagColor}{F6-argument}} & Neutral reporting of disputed legal issues or contentions advanced by the appellant or defendant (complaint, ground of appeal, application, purpose). \\
\textbf{F7} & \textbf{\textcolor{TagColor}{F7-jury}} & Statements of fact or view provided by the jury. \\
\textbf{F8} & \textbf{\textcolor{TagColor}{F8-other}} & Miscellaneous factual content not covered by other F sub-categories.\\
\textbf{F9} & \textbf{\textcolor{TagColor}{F9-admission}} & Defendant's admission, confession, or guilty plea, in either direct or indirect quote. \\
\textbf{F10} & \textbf{\textcolor{TagColor}{F10-assertion}} & Claims or statements by defendant(s), their counsels, or both sides, which \emph{may or may not be factual}. \\
\textbf{F11} & \textbf{\textcolor{TagColor}{F11-question}} & Questions asked or challenges raised during hearing or other court events. \\
\textbf{F12} & \textbf{\textcolor{TagColor}{F12-answer}} & Answers during hearing or other court events. \\
\textbf{F13} & \textbf{\textcolor{TagColor}{F13-objection}} & Objection of either side (usually to a question raised to the defendant) and relevant info (reason, sustained/overruled, outcome). \\
\textbf{F14} & \textbf{\textcolor{TagColor}{F14-instructions2jury}} & Judge's instructions given to the jury, in either direct or indirect quote. \\

\Xhline{1.4pt}

\rowcolor{InferColor}
\multicolumn{3}{l}{\textbf{I (Inference)} \, --- \, \textit{How does the court reason towards its decision?}} \\
\Xhline{1.4pt}

\textbf{I1} & \textbf{\textcolor{TagColor}{I1-case\_law}} & References to prior judicial precedents (common law) employed during reasoning. Includes content from previous judgments, in direct or indirect quote. \\
\textbf{I2} & \textbf{\textcolor{TagColor}{I2-ordinance}} & Citations of statutory laws, regulations, or ordinances used in the reasoning process. \\
\textbf{I3} & \textbf{\textcolor{TagColor}{I3-legislation}} & Citations of legislative documents, processes, organisations, or relevant info that \textbf{I2-ordinance} does not cover. \\
\textbf{I4} & \textbf{\textcolor{TagColor}{I4-conventional\_practice}} & Established customary practices (non-statutory) referenced during reasoning, such as reductions in sentencing (e.g.,\ one-third reduction). \\
\textbf{I5} & \textbf{\textcolor{TagColor}{I5-jury}} & Content related to the jury within the reasoning process. \\
\textbf{I6} & \textbf{\textcolor{TagColor}{I6-assertion}} & Judge's conclusive statement about the current case, such as assertion, concluding evaluation, result, etc., during inference.$^{\P}$ \\
\textbf{I7} & \textbf{\textcolor{TagColor}{I7-other}} & Miscellaneous inferential content not covered by other I sub-categories. \\
\textbf{I8} & \textbf{\textcolor{TagColor}{I8-question}} & Question raised by the judge as part of reasoning (vs.\ \textbf{F8-question} for questioning a party). \\

\Xhline{1.4pt}

\rowcolor{ResultColor}
\multicolumn{3}{l}{\textbf{R (Result)} \, --- \, \textit{What does the court decide?}} \\
\Xhline{1.4pt}

-- & \textbf{\textcolor{TagColor}{R}} & Final judgment determinations for a case. \\
-- & \textbf{\textcolor{TagColor}{R-other}} & Supplementary info adhered to a court determination (explanation, interpretation, clarification, calculation, consequence, effects). \\

\Xhline{1.4pt}

\rowcolor{OtherColor}
\multicolumn{3}{l}{\textbf{O (Others)}} \\
\Xhline{1.4pt}

-- & \textbf{\textcolor{TagColor}{O}} & Sentences that do not fit any of the above categories (e.g.,\ ``Court adjourns'', appearance records). \\

\bottomrule

\end{tabular}

\label{tab:schema}
\caption{The \textsc{HKJudge} sentence-level rhetorical role annotation scheme, comprising 26 tags grouped into four legal discourse functions: Fact (F), Inference (I), Result (R), and Other (O). The scheme was designed by legal linguistics experts and used as the reference during annotation.}
\end{table*}

\clearpage

\section{Example of Legal Linguistics Expert Annotation on Court of Final Appeal Judgment (FACC 22/2018)}
\label{sec:schema_exp}
\begin{figure*}[!ht]
    \centering
    \includegraphics[width=0.75\textwidth]{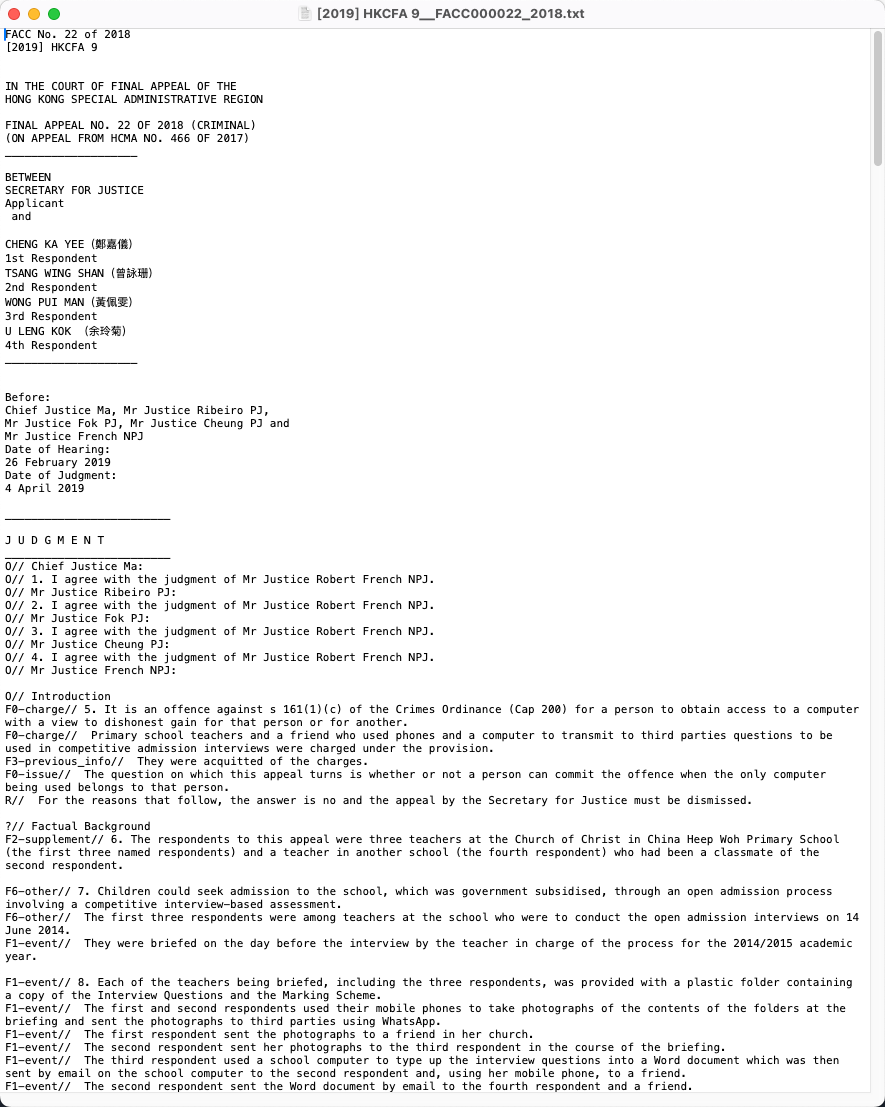}
    \caption{Example of sentence-level rhetorical role annotation in \textsc{HKJudge}, illustrated on a Court of Final Appeal judgment (FACC 22/2018). Each line is prefixed with its assigned rhetorical role tag.}

\label{fig:annotation_example}
\end{figure*}

\end{document}